\documentclass[letterpaper, 10 pt, conference]{ieeeconf}  

\IEEEoverridecommandlockouts                              

\usepackage{graphics} 
\usepackage{epsfig} 
\usepackage{amsmath}
\usepackage{amsfonts}
\usepackage{caption}
\usepackage{subcaption}
\usepackage{booktabs} 
\usepackage{algorithm}
\usepackage{algpseudocode}
\usepackage{dsfont}

\usepackage[dvipsnames]{xcolor}

\algnewcommand{\Initialize}[1]{%
  \State \textbf{Initialize:}
  \Statex \hspace{\algorithmicindent}\parbox[t]{.8\linewidth}{\raggedright #1}
}

\renewcommand{\vec}[1]{\mathbf{#1}}

\title{\LARGE \bf
Seq2Seq Imitation Learning \\ for Tactile Feedback-based Manipulation  
}

\author{Wenyan Yang$^{1}$, Alexandre Angleraud$^{1}$, Roel S. Pieters$^{1}$,  Joni Pajarinen$^{2}$, and Joni-Kristian K{\"a}m{\"a}r{\"a}inen$^{1}$
\thanks{$^{1}$Tampere University, Finland; $^{2}$Aalto University, Finland.}%
\thanks{2023 IEEE International Conference on Robotics and Automation (ICRA)}
}

\begin{document}

\maketitle
\thispagestyle{empty}
\pagestyle{empty}

\begin{abstract}

Robot control for tactile feedback based manipulation can be difficult due to modeling of physical contacts, partial observability of the environment, and noise in perception and control. This work focuses on solving partial observability of contact-rich manipulation tasks as a \textit{Sequence-to-Sequence (Seq2Seq)} Imitation Learning (IL) problem. The proposed Seq2Seq model first produces a robot-environment interaction sequence to estimate the partially observable environment state variables, and then, the observed interaction sequence is transformed to a control sequence for the task itself. The proposed Seq2Seq IL for tactile feedback based manipulation
is experimentally validated on a door-open task in a simulated environment and a snap-on insertion task with
a real robot. The model is able to learn both tasks from only 50 expert demonstrations while state-of-the-art reinforcement learning and imitation learning methods fail.


\end{abstract}

\section{INTRODUCTION}


The sense of touch is a key sensory modality for many robot manipulation tasks such as grasping~\cite{Bekiroglu-2011-ieeetr} and precision-insertion~\cite{Nakagaki-1995-icra,Yamamoto-2002-iros,Ma-2021-ieeetii}. Tactile sensing is an instrumental modality of robotic manipulation, as it provides information that is not accessible via remote sensors such as cameras or lidars. The key challenges in tactile sensing based control are the difficulty to accurately model physical contacts, partial observability of the environment from touch only, and noise in perception and control.
Tactile feedback based manipulation controllers have been proposed, but they often use heuristic search patterns~\cite{suarez2016framework} or are tailored for a specific task~\cite{Yamamoto-2002-iros}.
In such case, the learning-based approaches are more promising to learn generic solutions for contract-rich manipulation control tasks.


%
Reinforcement learning (RL) is one of the promising areas of machine learning for robotics. The goal of RL is to learn an optimal policy which maximizes the long-term cumulative rewards.
In the case of contact-rich manipulation, RL learning becomes challenging due to sparse reward and partial observability. That often results to an excessive amount of environment interactions needed, which is not doable
with real robots. An alternative option is to use simulations for teaching, but the difficulty to accurately model physical contacts becomes the bottleneck. Therefore, 
instead of pure RL, a more feasible solution is imitation learning (IL). In IL instead of trying to learn from the sparse rewards or manually specifying a reward function, an expert (typically a human) provides a set of demonstrations. The agent then tries to learn the optimal policy by following, imitating the expert’s decisions.
A number of imitation learning methods have been proposed~\cite{argall2009survey,calinon2007teacher,calinon2007incremental,billard2016learning,racca2016learning}, but these
do not particularly address the partial-observability problem that is inherent in tactile sensing.

In this work, we focus on solving contact-rich manipulation tasks with tactile-only sensing and in partially observable environments. Motivated by the above discussion, we aim to design a method that
safely and sample-efficiently learns contact-rich tasks with  minimal manual engineering. Safety in our case means that
IL following expert demonstrations better avoids generating dangerous actions than RL.
To address the partial observability, we take the common approach of RL for Partially-observable Markov Decision Processes (POMDPs): history data is used to aggregate belief of the partially observable environment states.
We combine it with IL framework to achieve sample efficiency. Two types of expert demonstration are used in this work: exploration and manipulation. The robot first imitates the expert's exploration for hidden state discovery, and use the exploration observations to produce a goal-directed trajectory that imitates the expert's  manipulation. The main contributions are:
\begin{itemize}
    \item A novel Sequence-to-Sequence (Seq2Seq) model to perform exploration-to-manipulation imitation learning. The model learns to translate the exploration trajectory into a manipulation trajectory.
    Generation of both types of trajectories is learned from expert demonstrations.    
    \item We show that a Transformer based Seq2Seq IL architecture is able to aggregate belief of hidden environment states during exploration. Besides, enforcing the Seq2Seq encoded feature to be similar to the hidden environment states contributes to task performance.
    \item The Proposed Seq2Seq IL is sample efficient. Sample efficiency is investigated and compared to strong baselines in the both simulated and real manipulation tasks.
    Seq2Seq IL learns successful control policies from only 50 expert demonstrations.
\end{itemize}
%


\section{Related work}
The existing controllers for tactile feedback manipulation often use control heuristics and need a model for the contact dynamics~\cite{Yamamoto-2002-iros,suarez2016framework, johannsmeier2019framework}.
Recently, a number of methods combining force kinematics with compliant control and machine learning have been proposed to overcome the need of manual tuning~\cite{ranjbar2021residual, vuong2021learning,dong2020compliance,kutsuzawa2018sequence,si2022adaptive}. Moreover, Imitation Learning (IL), also known as Learning from Demonstration (LfD), allows more flexible “programming” of a robot as it learns from expert demonstrations. Various computational models, including Gaussian Mixture Models, Hidden Markov Models, Deep Neural Networks and Dynamic Motion Primitives, have been proposed for IL~\cite{argall2009survey,calinon2007teacher,calinon2007incremental,billard2016learning,racca2016learning}. These approaches differ from our work rather strongly in the sense that they assume a fully observable environment or a model for the contact physics or are task specific ("Peg-in-Hole" or "goal reaching"). We seek for a generic method that does not need to model the contact physics.

A number of generic RL and RL-based IL methods exist~\cite{xu2018feedback,zhao2021offline,spector2021learning,khader2020stability,levine2014learning,apolinarska2021robotic,spector2020deep,luo2018deep}. However, they as well assume a fully observable environment that can be modeled as a Markov Decision Process (MDP), and require a large number of interaction steps to converge. The partial-observability problem (POMDP) has been addressed in a number of works~\cite{bagnell2001solving,kwon2020inverse,gangwani2020learning,arjona2019rudder,ren2021learning,meng2021memory,yang2021recurrent, han2019variational,singh2021structured}. We adopt the common approach in these works, and use the observation history to aggregate belief of the
partially observable state variables. The aggregation can be done by sequential
learning using recurrent neural networks~\cite{Hausknecht-2015-AAAI,Igl2018icml,Lee-2020-neurips,meng2021memory,igl2018deep}. The main shortcoming of the above  RL approaches for POMDPs is that they need a massive amount of environment interactions which makes them unsuitable for real robot tasks.

From the imitation learning perspective, a numerous work have proposed to learn a model-free policy for expert demonstrations tasks~\cite{GAIL,IL1,IL2,IL3,IL4}. However, these methods still requiures online data sampling and is not practically sample-efficient for real applications. Some offline imitation learning methods such as implicit behaviour cloning (IBC)~\cite{IBC}, DemoDice~\cite{demodice}, etc., but they are not designed for solving POMDP problems.

\section{Background} 
\label{background}

%
%
\subsection{Tactile feedback in manipulation}
\label{sec:tactile_feedback}

\begin{figure}[bh]
\includegraphics[width=1.\linewidth]{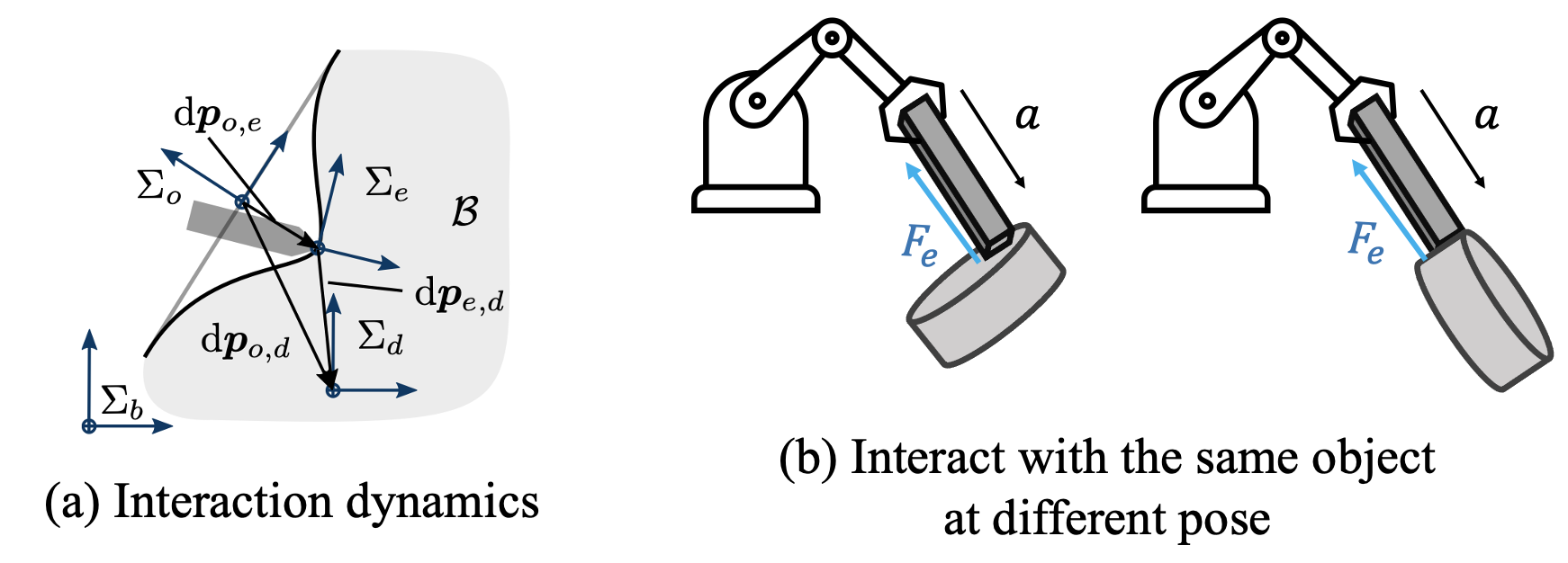}
\caption{Illustration of the physical wrench (torques and forces) during robot manipulation: (a) the interaction frames involved in Eq.~\ref{eq:interaction_dynamics} and (b) an example where the observed wrench is the same for two distinct surface (task) points.}
\label{fig:dynamics}
\vspace{-5pt}
\end{figure}

In this work, the contact sensory feedback refers to the external wrench $\textbf{F}_{ext}$ observed at the robot end-effector. Tactile feedback provides only partial observation of the environments, the object pose in our example. This can be analytically studied via the contact kinematics investigated in mechanical engineering and robotics~\cite{garg2019learning,Montana-1988-ijrr,Howard-1998-mmt}.

Consider a robot-environment interaction model, where the robot interacts with a target surface directly with its end-effector or through an object in its gripper. Suppose the environment is  defined by its stiffness, and the end-effector is treated as a rigid object. For rigid objects the roles can be interchanged.
Denote $\Sigma_b$ as the Cartesian base frame, $\Sigma_d$ is the target object frame, $\Sigma_e$ is the end-effector frame,  and $\Sigma_o$ is the task frame  located on the target object surface. In this setting, the surface provides resistance to the robot end-effector's attempts to penetrate the target. This resistance can be modeled as the external wrench~\cite{Gold-2020-iros}
\begin{equation} 
{
^o\mathcal{\textbf{F}}_{ext} 
 = ({^o\textbf{K}_e} + {\textbf{K}_{P, cart})}^{-1} {^o\textbf{K}_e} {\textbf{K}_{P, cart}} \hbox{d}{\textbf{p}_{o,d}}} \enspace ,
\label{eq:interaction_dynamics}
\end{equation} 
%
where the overall stiffness matrix consists of a Cartesian controller stiffness matrix $\textbf{K}_{P, cart}$ and the environment's resistance-to-penetration stiffness matrix $^o\textbf{K}_e$. $^o\textbf{K}_e$ is composed by the translational and rotational components, but here we omit the rotation component and assume purely translational stiffness matrices for simplicity.

The translational interaction is given by $\hbox{d}\textbf{p}_{o,d}$ 
that is the displacement of the task frame $\Sigma_o$ with respect
to the target frame $\Sigma_d$ (Figure~\ref{fig:dynamics}(a)). The two terms for the analysis are $^o\textbf{K}_e$ that depends on the target object's material and $d\textbf{p}_{o,d}$ that depends on the normal of the object's surface at the task frame $\Sigma_o$.
%
%
These terms verify that the observable wrench in Eq.~\ref{eq:interaction_dynamics} depends on the surface stiffness matrix (end-effector assumed rigid) and on the surface normals with respect to the end-effector frame $\Sigma_e$ as illustrated in Fig.~\ref{fig:dynamics}(a). If we assume homogeneous stiffness properties for the target object that means that the observed wrench is the same for all task frames, surface points $\Sigma_o$ for which the end-effector-surface normals are equal. This is illustrated for a solid object in Fig.~\ref{fig:dynamics}(b). \textit{To summarize, it is not possible to infer the object pose from a single touch sensor measurement.}

%
%
\subsection{Partially Observable Markov Decision Process}
If the underlying environment state cannot be fully ascertain, the problem can be formulated as POMDP 
~\cite{kaelbling1998planning}. Formally, a POMDP can be described as a 6-tuple $(\mathcal{S},\mathcal{A},\mathcal{P},
\mathcal{R},\Omega,\mathcal{O})$ where $\mathcal{S}$, $\mathcal{A}$, $\mathcal{P}$, $\mathcal{R}$, are the states,
actions, transitions and rewards. At time $t$, the environment is in some state $s_t \in S$, and agent generates
action $a_t \in A$. The environment produces a new  state $s_{t+1} \in S$ based on dynamics $T(s_{t+1}|s_t,a_t)$ and the
agent receives a reward $r_t = R(s_t , a_t , s_{t+1})$. However, the agent cannot directly observe the underlying state
$S$ in a POMDP. Instead, the agent receives an observation $o_t \in O$  via the indirect observation function $O(s_{t+1},
a_t, o_t) = P(o_t | s_{t+1},a_t)$. In general, this implies that the agent must take the entire history of observations
and actions $h_t = ((a_0,o_0), (a_1, o_1), \ldots, (a_{t-1}, o_{t-1}))$ into account to 
make the current state more observable~\cite{nguyen2020belief}.

Based on our analysis of the tactile feedback, the uncertainty of the target object pose from touch sensing makes the task environment only \textit{partially observable} for a force-feedback controller. However, the history of tactile feedback helps to estimate the pose which eventually helps to solve the manipulation task.

\subsection{Tactile-only manipulation tasks}

\begin{figure*}[t]
\centering
\includegraphics[width=.99\linewidth]{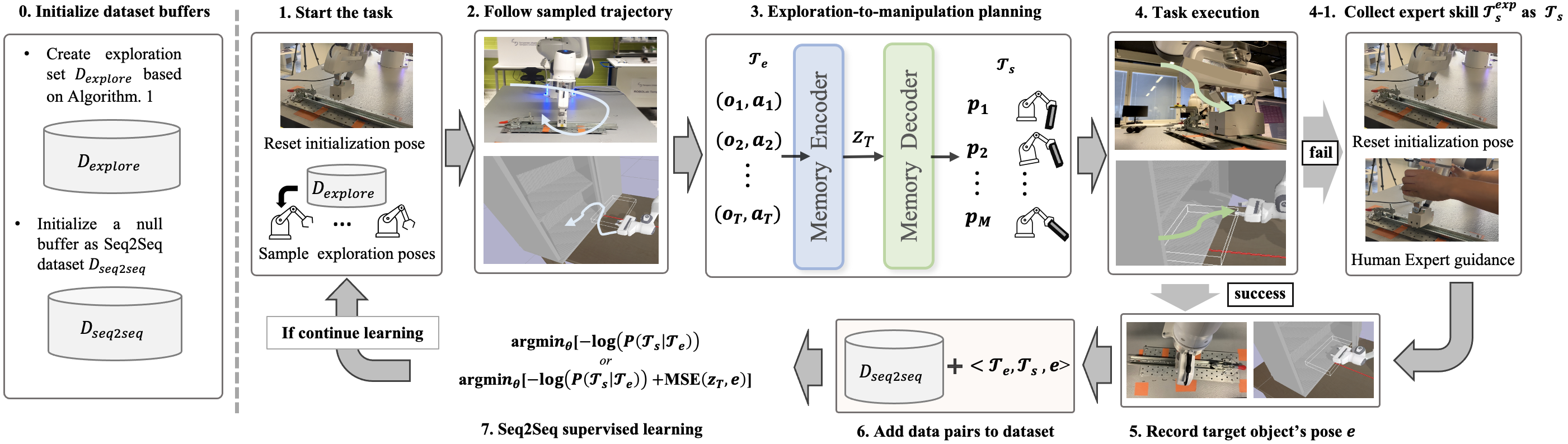}

\caption{The proposed Seq2Seq Imitation Learning pipeline. \textbf{Step 0}: initialize the datasets. \textbf{Step 1-2}: the robot first samples and executes one of the given expert exploration trajectories to collect an observation trajectory $\mathcal{T}_{e}$. \textbf{Step 3-4}: the encoder infers an underlying environment state $z$ from $\mathcal{T}_{e}$, and then the decoder plans and execute  a skill trajectory $\mathcal{T}_{s}$. If the task fails, \textbf{Step 4-1}: the robot returns to initial pose and expert guides the robot to complete the task, meanwhile record expert skill trajectory $\mathcal{T}_{s}^{exp}$. \textbf{Step 6-7}:. the sequence pair $\{\mathcal{T}_{e}, \mathcal{T}_{s}, e\}$  are incrementally added to dataset and are used to fine-tune the imitation model. The pipeline stops if the imitation model successfully accomplish the task with 10 continuous tests.} \label{setup}
\vspace{-10pt}
\label{fig:pipeline}
\end{figure*}

\begin{figure}[h]
\includegraphics[width=0.99\linewidth]{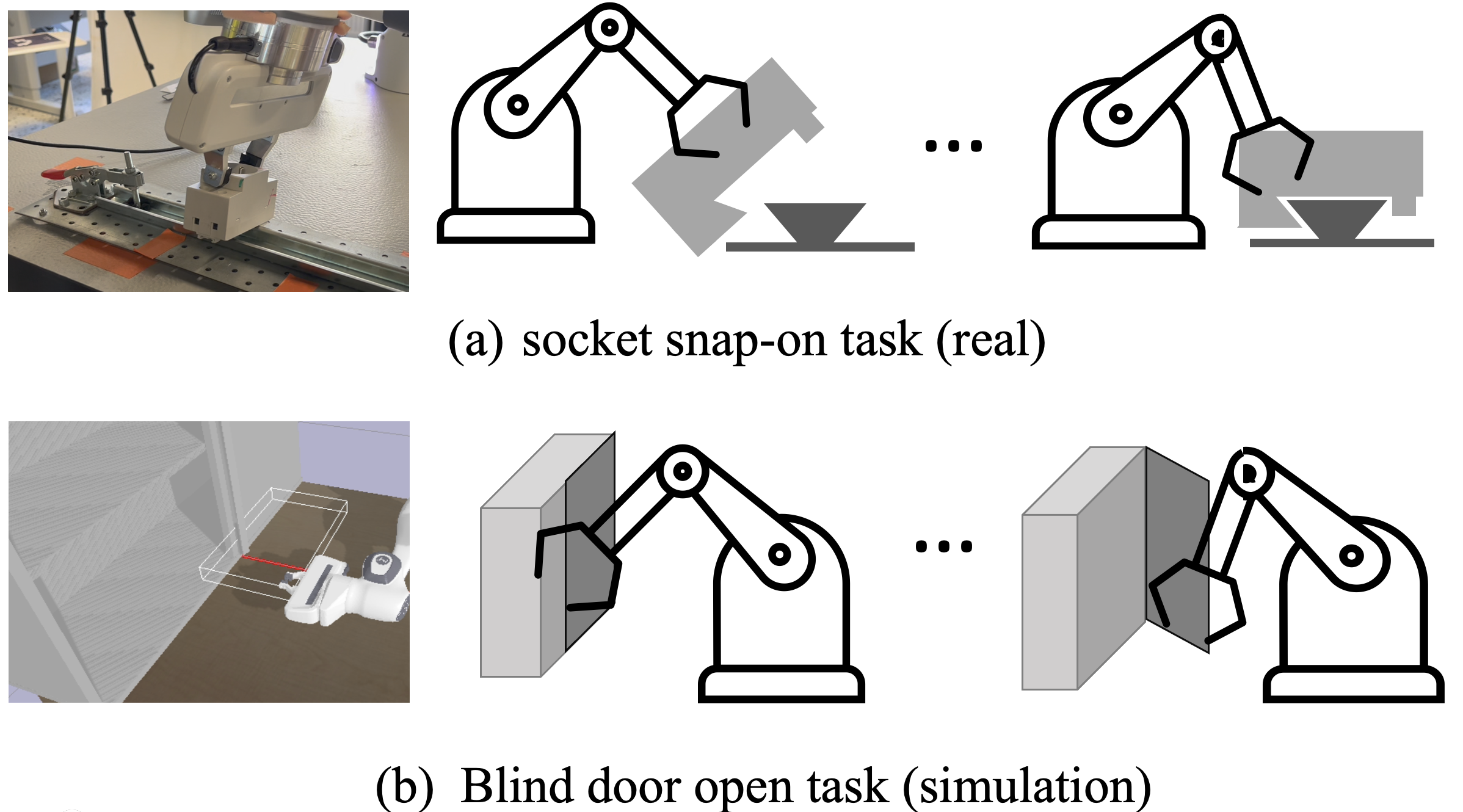}
\caption{The real (a) and simulated (b) test environments. The relative pose of the installed socket and the assembly rail (a) and the robot-end effector and the cabin door (b) are randomized in the experiments.}
\label{fig:two_tasks}
\vspace{-10pt}
\end{figure}
In this work, two high-precision manipulation tasks are studied as representations of tactile-only manipulation tasks. The two test environments, as shown in Figure~\ref{fig:two_tasks}, consist of a simulated door-opening task and a real snap-on insertion task:  \paragraph{Simulated door-open task} This task is a customized version of the "Opening" task in the MiniTouch benchmark in~\cite{rajeswar2022haptics}
and implemented in PyBullet. In our task the cabin
is randomly placed in the front of a virtual Franka Emika Panda.
The task is to open the cabin door using external wrench feedback and robot's states (pose and velocity of the Emika end-effector).

\paragraph{Snap-on mounting task}  The real snap-on mounting is a high-precision assembly task where Panda Emika needs to mount an electronic socket into an assempy rail (see Fig.~\ref{fig:two_tasks}). Similar to the above simulated task the robot uses only end-effector's external wrench feedback and end-effector's pose and velocity.

 To complete these tasks, the robot must accurately estimate the relative pose of the end-effector with respect to the manipulated object (such as the door edge or assembly rail).
The tasks are particularly challenging for tactile-only sensing for two reasons: first, these tasks require high precision, with the snap-on task requiring rotations smaller than one degree and distances smaller than five millimeters, and the door-opening task requiring a relative distance smaller than one centimeter. Second, the relative poses cannot be directly measured by a tactile sensor (as outlined in Section~\ref{sec:tactilefeedback}). To successfully complete the manipulation tasks, the robot must explore the environment in order to observe and better understand its state.


\section{Seq2Seq Imitation Learning}
\label{sec:Seq2Seq_model}

Imagine manipulating objects in a dark room. For example, find and pick up your coffee cup. After finding the cup, but before grasping it, you do "tactile exploration" to find its handle and to adjust your grip. Inspired by this strategy, we introduce a model and learning procedures for Sequence-to-Sequence imitation learning (Seq2Seq IL). The robot first executes an exploration trajectory to collect and infer the hidden information of the environment, then it plans a trajectory according to the aggregation of the collected information.

\begin{figure}[t]
\centering
\includegraphics[width=0.95\linewidth]{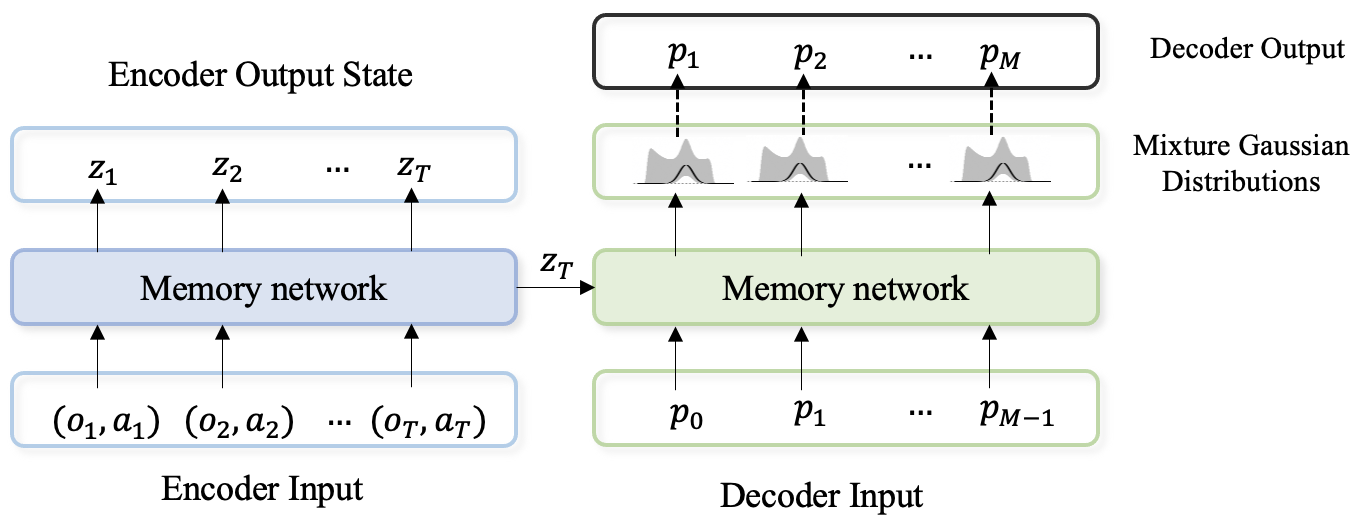}\caption{The overall Seq2Seq architecture used in our model. The input sequence is an exploration trajectory $\mathcal{T}_{e}=\left\{(o_t,a_t)\right\}$ and the output sequence is a skill execution trajectory $\mathcal{T}_{s}=\left\{p_m\right\}$. The encoder performs feature aggregation to infer the current state $z_T$ of a partially observable environment.}
\label{fig:transformer}
\vspace{-10pt}
\end{figure}

Sequence-to-sequence is a common approach to solve sequential problems~\cite{sutskever2014sequence}. The problem can be defined as sequential mapping of one $T$-length input sequence $X=\{x_1, x_2, ..., x_T\}$ to an $M$-length output sequence $Y=\{y_1, y_2, ..., y_M\}$.
The overall structure of our Seq2Seq model is 
illustrated in Fig.~\ref{fig:transformer}. The model contains two memory modules which are linked by an encoder-decoder structure. The encoder sequentially receives inputs which are encoded to an internal representation $z_t$. In our terminology, the encoder performs the environment exploration and produces $z_t$ as a state estimate of partially observable environment. The skill planning is performed by the decoder that auto-regressively generates the target sequence after receiving the encoder's state estimate $z_t$. The Seq2Seq model can be implemented as an LSTM network~\cite{sutskever2014sequence}, or more recently, as a 
Transformer network~\cite{brown2020language}.

As a distinct feature to other similar works, our Seq2Seq model breaks the tactile feedback manipulation tasks into two stages: 1) \textit{exploration stage} and 2) \textit{skill planning stage}.
In the exploration stage the robot follows the encoder trajectory to explore the environment and encode its state into the internal representation $z_t$. In the skill planning stage, the decoder produces a goal-directed control trajectory for the low level robot controller to complete the main manipulation task. The both encoder and decoder trajectories are learned via imitation learning, i.e. from expert demonstrations. 

\subsection{Sequence modeling} 
For the exploration stage, a recorded expert exploration control trajectory (via-points) are given to the robot. The robot follows the expert trajectory to collect tactile feedback from
the environment. The usage of expert trajectories and the robot in impedance control mode allows safe exploration. $\mathcal{T}_{e}$ is the observed $T$-length exploration trajectory,
\begin{equation}
    \mathcal{T}_{e} = \{(o_1,a_1), (o_2,a_2),..., (o_{N}, a_T)\} \enspace ,
\end{equation}
where $o_i$ is an observation containing 18 attributes: end-effector external wrench $^o\textbf{F}_{ext}$ (translational and rotational components), end-effector pose (3D translation and orientation), and its velocity (translational and rotational).
$a$ is a $6$-dimensional action vector of the end-effector's displacement.
The skill planning trajectory (via-points) $\mathcal{T}_{s}$ is
\begin{equation}
    \mathcal{T}_{s} = \{p_{1}, p_2, \ldots, p_M\} \enspace ,
\end{equation}
where $p_i$ is a 6-dimensional end-effector pose.

Given the above definitions of the exploration and skill planning trajectories, the problem can be cast as a Seq2Seq problem. A Seq2Seq controller provides the robot low level controller a task-directed trajectory (a via-point sequence) $\mathcal{T}_{s}$ from the exploration (and observation) sequence $\mathcal{T}_{e}$.

\subsection{Seq2Seq encoder-decoder} 

\subsubsection{Exploration encoder}
In our encoder-decoder structure, the Seq2Seq encoder gradually observes $\mathcal{T}_{e}$ and encodes the observations into an internal representation $z_T$. In our formalism we assume that the encoder representation includes the state variables that are only partially observable, but for which belief is aggregated through history data. The encoder is depicted in the left-hand-side of Fig.~\ref{fig:transformer} and is formally
\begin{equation}
    z_T=\hbox{Enc}(\mathcal{T}_{e})
\end{equation}
where $z_T$ is the encoded representation and approximates belief state of the POMDP.

\subsubsection{Skill planning decoder}
The Seq2Seq decoder receives $z_T$ from the encoder and produces the skill plan trajectory $\mathcal{T}_{s}$. The history of the previous planned poses is $\tau_{t-1}=\{p_0, ..., p_{t-1}\}$.
The probability of the next skill planning pose is $P(p_t|z_T, \tau_{t-1})$, and is modeled as a mixture of $K$ Gaussians. The Gaussian mixture probability of planned pose $p_t$ at the step $t$ is defined as
\begin{equation}
     P(p_t|z_T,\vec{\tau}_{t-1}) = \sum^K_{i=1} w^i_\theta\mathcal{N}(\mu^i_ \theta, \sigma^2_{i,\theta}) 
    \label{eq:mdn_loss}
\end{equation}
where $w^i_\theta$ is the weight of the $i$-th Gaussian, $\mu^i_\theta$ and $\sigma^2_{i,\theta}$ are the mean and variance of the $i$-th Gaussian. The parameter $\theta$ denotes that these are estimated by a Mixture Density Network (MDN) layer applied at the decoder's head (Fig.~\ref{fig:transformer}).
The Seq2Seq learning objective is to maximize the probability of a target sequence, or respectively, minimize the negative log-likelihood.  
The loss can be formulated as:

\begin{equation}
    \mathcal{L}_{seq2seq} = -\log P(\mathcal{T}_s|\mathcal{T}_e) = -\sum^M_{t=1} \log P(p_t|z_T, \tau_{t-1}) \enspace .
\label{eq:objective_Seq2Seq}
\end{equation}

\subsubsection{Supervised Seq2Seq}
If the partially observable state variables are known; let's denote
them by $e$; then these can be embedded to the encoder output $z_T$ and
learned supervised manner.
A suitable loss is, for example, the standard mean-squared error (MSE) loss.
The supervised Seq2Seq loss is
%
\begin{equation}
    \mathcal{L}_{supervised} = \mathcal{L}_{seq2seq} +||z_T-e||^2 \enspace .
\label{eq:explicit_loss}
\end{equation}
%
The supervised Seq2Seq IL and 
Seq2Seq IL are two variants used in our experiments.
We denote the supervised
version as "Seq2Seq-oracle" as, in general, the partially observable variables are not known.
\section{Experiments}

\subsection{Settings}
\label{sec:implementations}

\subsubsection{Demonstrations collection and imitation pipeline} Since our model consists of two kind of trajectories, exploration and skill planning, a small number of initial demonstrations of the exploration are first provided, and then the actual expert skill execution demonstrations are demonstrated in the human-in-the-loop manner. Figure~\ref{fig:pipeline} illustrates the complete training procedure.

For exploration, a human expert
provides a small number (five in our case) exploration trajectories. The experts were instructed to move the robot hand in the kinesthetic teaching mode such that they "feel" that the socket is in the correct position with respect to the rail.
See Algorithm~\ref{alg:explore} for details. 

After the initial exploration demonstrations the target (rail/door) pose is randomized. The robot produces and executes an exploration trajectory and given the exploration feedback ($\mathcal{T}_e$) produces and executes a skill planning trajectory ($\mathcal{T}_s$). If the task is successful, the both trajectories ($\mathcal{T}_e$ and $\mathcal{T}_s$) are added to the training data. If the task fails, the robot position is reversed
back to the position just before the skill execution, and an expert provides a successful skill execution $\mathcal{T}_{s}^{exp}$. Then ($\mathcal{T}_e$ , $\mathcal{T}_s^{exp}$) will be added to training data. The most important expert demonstrations for our method are these human-in-the-loop failure correction demonstrations. We apply DAgger-like style learning~\cite{ross2011reduction} to incrementally collect data to learn from the experts (see Fig.~\ref{fig:pipeline}). Our supplementary material contains video clips about training and testing.

For the real snap-on task all demonstrations are provided manually by a human expert. For the simulated door-open task the exploration trajectories are provided manually by an expert (mouse is used to move the end-effector to the door and then move it on the door surface), and the skill execution demonstrations are provided by a SAC trained controller that fully observes the environment (the cabin 3D pose).

\begin{algorithm}
     \caption{Expert exploration trajectories collection}
     \label{alg:explore}
    \begin{algorithmic}[1]
        \State Initialize expert exploration set $D_{exp}$. skill set $D_{s}$. Denote the robot end-effector pose as $o$.
        \For{$i=1$ to $n$}
            \State Initialize robot to its starting pose\;
            \State Set robot to the kinesthetic teaching mode\;
            \State Randomize target object pose 
            \State Let expert guide end-effector to explore\;
            \State Record and store expert trajectory $\tau_{explore} = \left\{\vec{o}_0,\vec{o}_1,\ldots,\vec{o}_T\right\}$, where $o_t$ is the end-effector pose\;
            \State Add $\tau_{explore}$ to $D_{exp}$\;
        \EndFor
    \end{algorithmic} 
\end{algorithm}

\subsubsection{Baseline methods and metrics}
We use the following criteria to select the baseline methods: 1) the baseline shall be learning-based methods and 2) the baseline shall be designed to solve the POMDP problems. Based on these criteria, we chose the following  baseline methods from multiple method categories.
\textbf{POMDP Reinforcement Learning (POMDP-RL):}
\begin{itemize}
\item Recurrent model-free RL using SAC or TD3~\cite{ni2021recurrent}: RMF-RL(sac) and RMF-RL(td3)
\item RMF-RL modified by adding the Behavior Cloning (BC)~\cite{nair2018overcoming}: RMF-RL-IL(sac) and  RMF-RL-IL(td3)
\item Soft-Actor-Critic~\cite{haarnoja2018soft} that observes the full environment (with "oracle"): SAC-MDP
\end{itemize}
\textbf{Imitation Learning (IL):}
\begin{itemize}
\item Soft-Q Imitation Learning is a RL-based IL which assigns sparse rewards to expert demonstrations~\cite{reddy2019sqil} (due to the POMDP setting, the RL part is RMF-RL(sac)): SQIL
\item A classical behavior cloning (BC)~\cite{bain1995framework} where an LSTM network is used to modernize it: BC-LSTM
\end{itemize}
\textbf{Our Seq2Seq IL variants:}
\begin{itemize}
\item Seq2Seq - a Transformer Seq2Seq IL that learns the partially observable state variables through exploration ($\mathcal{L}_{seq2seq}$  in Eq.~\ref{eq:objective_Seq2Seq})
\item Seq2Seq-oracle - a Transformer Seq2Seq IL that is trained supervised manner ($\mathcal{L}_{supervised}$  in Eq.~\ref{eq:explicit_loss})
\item Seq2Seq-LSTM - Otherwise similar to Seq2Seq IL, but the Transfomer is replaced by LSTM from~\cite{Hausknecht-2015-AAAI}
\end{itemize}
In all experiments, the length of the encoder output $z_T$ is set to 3 as it corresponds to the number of actual partially observable task variables, the xy-place location of the target and its rotation.

\subsubsection{Success rate}
Success rate is used as the  performance metric.
Success rate reports the proportion of test runs where the agent reached the goal state (door opened / socket mounted). 

\begin{figure}
    \centering
    \includegraphics[width=1\linewidth]{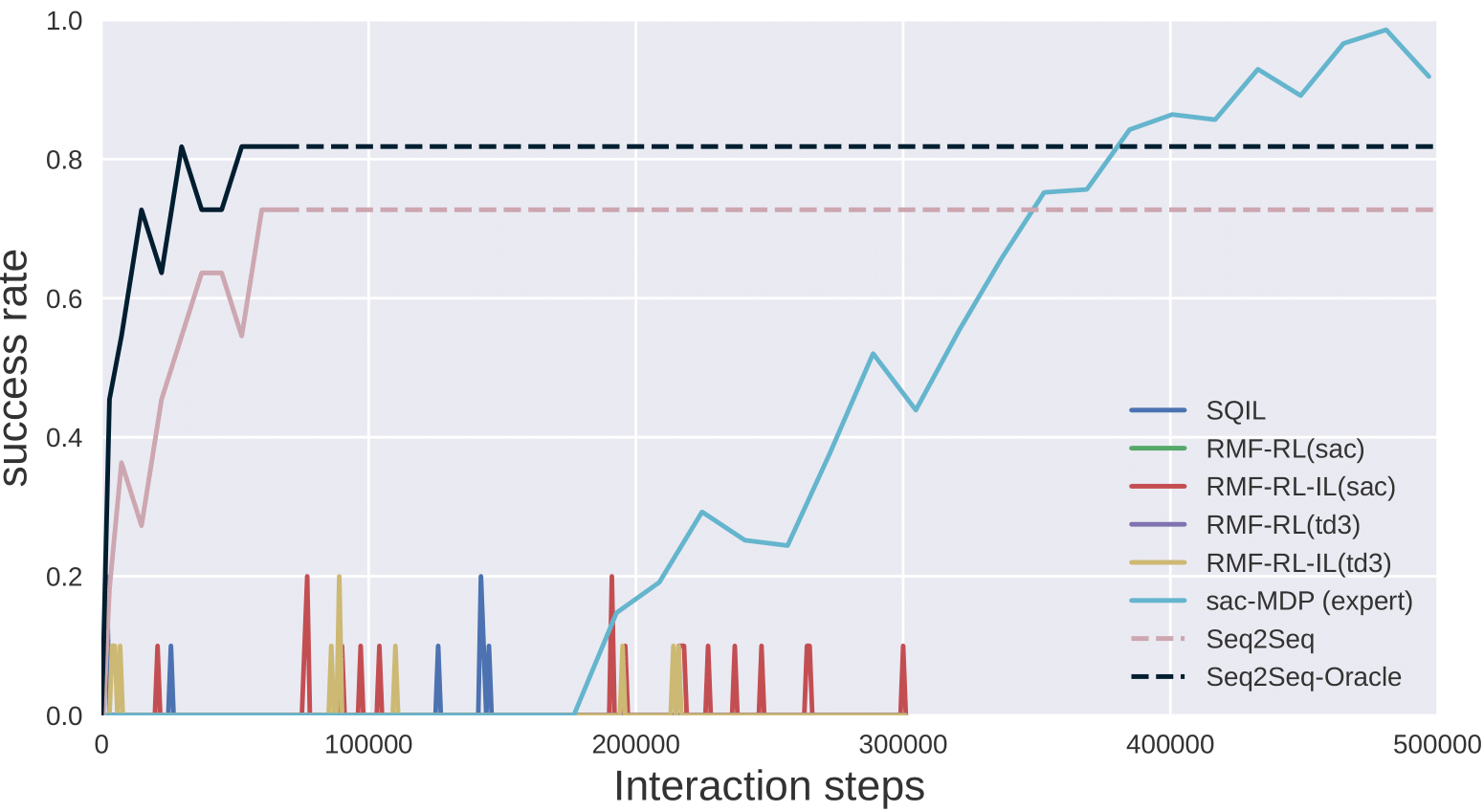}
    \caption{Sample efficiency experiment (simulated door-open task): success rates as the function of interaction steps. Our methods were trained only for 50 demonstrations (approx. 67k interactions).}
    \label{fig:sample_effciency}  
\vspace{-15pt}
\end{figure}

\subsection{Results}

\subsubsection{Sample efficiency}
Since sample efficiency is one of the main limiting factor in using machine
learning techniques in real robot learning tasks, the first experiment evaluates
sample efficiency of the proposed model and compares it to the POMDP RL and IL baselines. The experiment was conducted using the simulated door-open task
to allow a large number of samples and to avoid failures that would damage the real robot. 50 expert demonstrations were given to RMF-RL-IL and SQIL. Seq2Seq and Seq2Seq-Oracle were trained using DAgger style and stopped after 50 demonstrations. Figure~\ref{fig:sample_effciency} presents the results.


We used an oracle trained soft-actor-critic (SAC-MDP) as the expert. The oracle SAC-MDP agent observes the door pose, robot's end-effector external wrench, pose, and velocity. SAC-MDP successfully learns the task, but requires more than 400k interactions. It is worth noting that none of the POMDP RL baselines (RMF-RL(sac/td3)) was able to solve the problem even after 500k interactions. The imitation RL methods (SQIL and RMF-RL-IL(sac/td3)) achieved 20\% success rate, while their performance was unstable. Using DAgger-style incremental learning, our proposed model (Seq2Seq) achieved 76\% success rate. 

Our Seq2Seq models were the only to learn the task without oracle (POMDP setting) and required only 50 demonstrations that corresponds to approximately 67k interaction steps. This is clearly better than the oracle
SAC-MDP that had fully observable environment. The supervised variant of our method, Seq2Seq-Oracle, obtained 16\% higher success rate (82\%) than the POMDP Seq2Seq. This indicates that the model can effectively discover the partially observable state variables even without supervision.

\begin{table}[b]
\vspace{-10pt}
\caption{Imitation learning method comparison.}
\centering
\resizebox{0.99\columnwidth}{!}{
  \begin{tabular}{lccccccc}
    \toprule
    \textbf{Env.} &

      \multicolumn{1}{c}{Exp.$^\star$}&
      \multicolumn{1}{c}{SQIL} &
      \multicolumn{1}{c}{BC-} &
      \multicolumn{1}{c}{Seq2Seq-}&
      \multicolumn{1}{c}{Seq2Seq} &
      \multicolumn{1}{c}{Seq2Seq-}  \\
      & & &LSTM & LSTM &  & Oracle \\
    \midrule
Door-open &  $95\%$ &  $20\%$ & $11\%$   & $56\%$   &$76\%$ & $89\%$ \\
Snap-on    &  $100\%$  & - & $5\%$   & $0\%$ & $84\%$ & $88\%$\\
    \bottomrule
  \end{tabular}}
  
  $^\star$ 500k trained SAC for door-open and a human expert for snap-on
\label{tab:method_comparison}
\end{table}

\subsubsection{Performance evaluation}


We further evaluated IL methods' performance (success rates). 50 demonstrations were used to train each of them.
The results are in Table~\ref{tab:method_comparison} (100 random tests for each).

\begin{figure}[h]
    \centering
    \includegraphics[width=0.99\linewidth]{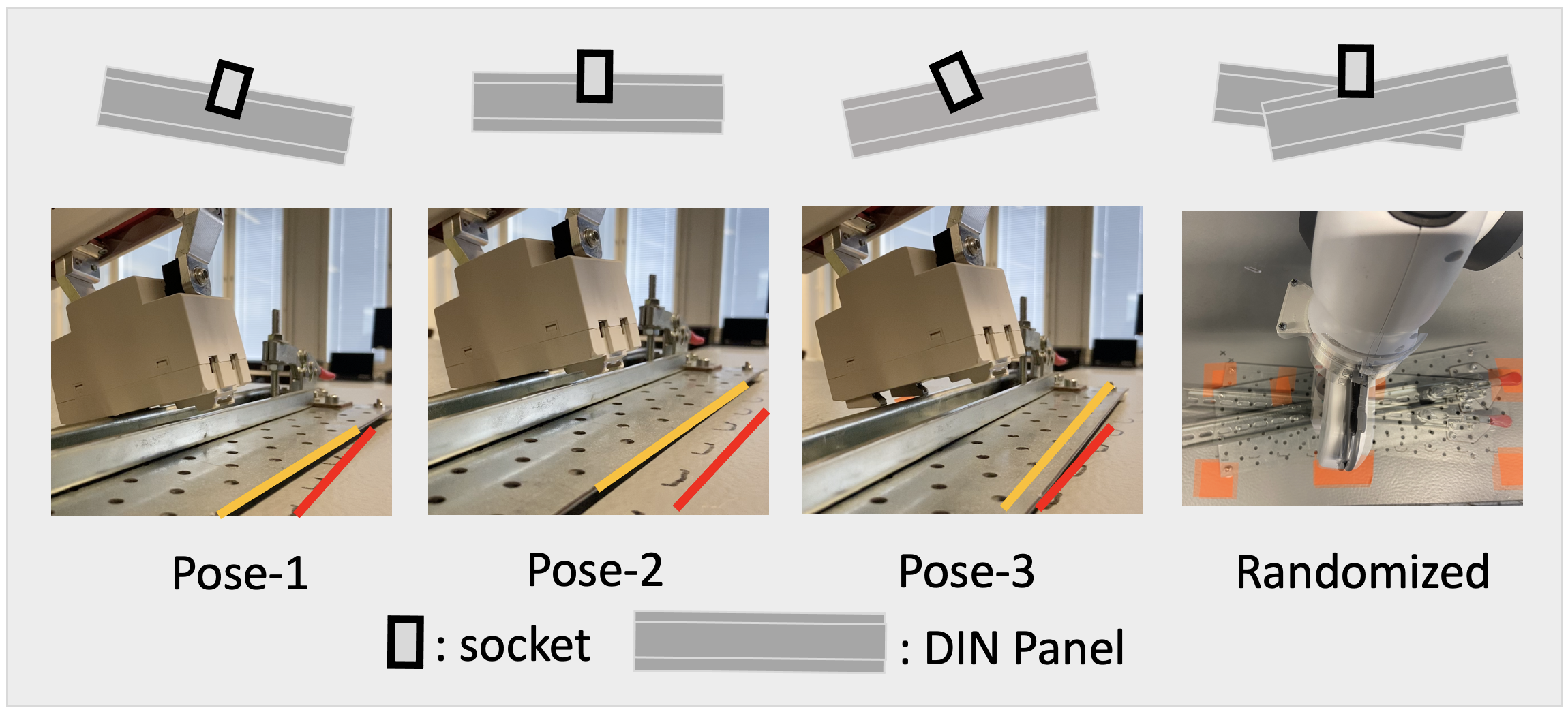}
    \caption{Examples of the poses used in the robustness and repeatability experiment (the real setup)} \label{fig:randomize_setup}
\end{figure}

\vspace{-10pt}

\begin{table}[t]
\caption{Repeatability/robustness results for the real robot task (snap-on)}
\begin{center}
\resizebox{.95\linewidth}{!}{
  \begin{tabular}{lcccc}
    \toprule
     {} &
      {Randomized}&
      {Pose-1} &
      {Pose-2} &
      {Pose-3}\\
    \midrule
    Success & ${84/102}$  & ${20/20}$&  ${20/20}$ &  ${18/20}$  \\
    \bottomrule
  \end{tabular}}
\end{center}
\label{tab:real_success_rate}
\end{table}

SQIL failed on the snap-on task with the real robot and achieved only
20\% success rate in the simulated door-open task. BC-LSTM also had poor performance on the both tasks. Seq2Seq-LSTM performed better than the plain BC-LSTM on the simulated door-open task, but completely failed on the real snap-on task. The Seq2Seq model has the second-best performance, 76\% success rate on the simulated and 84\% on the real snap-on task. This verifies that 1) the Seq2Seq imitation model can learn POMDP tasks effectively, and 2) the Transfomer-based Seq2Seq imitation model is more robust across the tasks than LSTM-based. Our proposed Seq2Seq-Oracle obtained the best performance in both tasks.

\subsubsection{Detailed analysis on the real robot}
We selected the Seq2Seq-Oracle to conduct a more detailed experiments
on the snap-on task with a real robot.

\paragraph{Robustness to partially observable state variables} 
To evaluate the robustness of the proposed model we conducted
repeatability and robustness experiments. For the repeatability
test we executed the learned model 20 times for the rail in
a fixed position (Pose-1, Pose-2 and Pose-3, see Fig.~\ref{fig:randomize_setup}). To evaluate robustness of
the estimation of the partially observable state variables
(pose of the rail), we executed the model 102 times for
random poses.

The results for the repeatability and robustness experiments are in Table~\ref{tab:real_success_rate}. The Seq2Seq IL model succeeded
84 out of the 102 attempts with the random rail poses. This
indicates fairly good robustness of the method to estimate the
non-directly observable state variables. For the two fixed poses, Pose-1 and Pose-2,
the repeatability was 100\% and 90\% (18/20) for Pose-3. These results indicate that the model learns to complete the task, is robust
to environment changes and insensitive to observation noise.

\begin{figure}[h]
\centering
\includegraphics[width=1.0\linewidth]{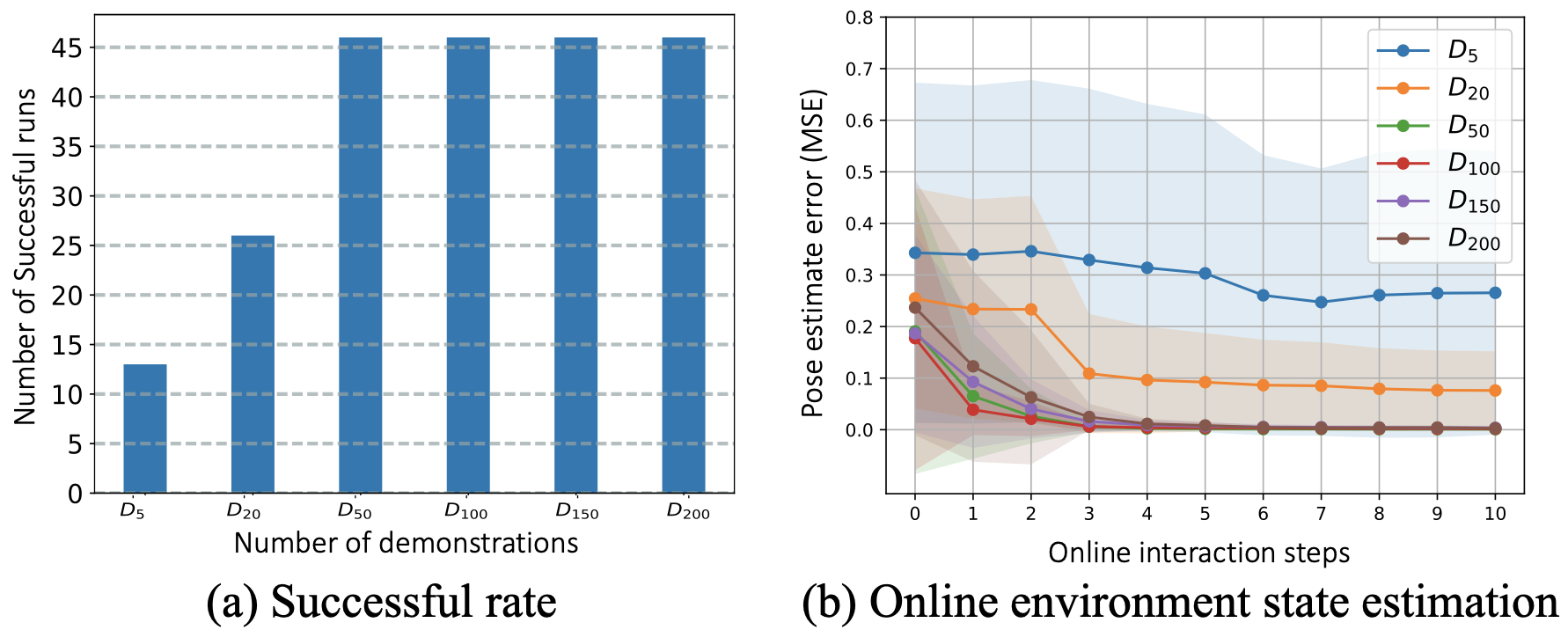}
\captionof{figure}{(a) number of successful test attempts for the model trained with different number of expert demonstrations and (b) Online error in the partially observable state variable estimation (pose) for the same models.}
\label{fig:dagger_performance}
\vspace{-15pt}
\end{figure}

\paragraph{Number of expert demonstrations} In order to test how effectively the model learns from new expert demonstrations, we recorded 200 expert demonstrations and
randomly formed training sets of 5 to 200 demonstrations. These training sets were used
to train the model and then the model was tested 45 times.
The results are shown in Fig.~\ref{fig:dagger_performance}(a). The results verify that the performance rapidly improves and converges after 50 expert demonstrations.


\paragraph{Online state estimation} In the final experiment we investigated the exploration
part of the Seq2Seq model. Using Seq2Seq-oracle we can obtain current estimates of the partially observable state variables during exploration. In the snap-on task, the variables are the pose of the rail defined by xy-plane coordinates and the rotation angle. We normalized translations and angles to the approximately same scale and computed the error (MSE) online after each exploration step. The results are in 
Fig.~\ref{fig:dagger_performance}(b) and verify that only 4 exploration steps are needed to obtain an accurate estimate of the state variables. This result holds if at least 50 expert demonstrations ($D_{\ge 50}$) are used to train the model. 

\section{Conclusions}
This work investigated the partial observability problem in learning control policy for tactile-feedback based manipulation. We proposed a transformer-based \textit{Seq2Seq} imitation learning (IL) which imitates expert exploration trajectories, and from them plans a suitable skill trajectory to
complete the task. The two stages of Seq2Seq, exploration and skill planning, are learned from expert demonstrations. The proposed model
is sample efficient and learns to solve a real snap-on task from only 50 expert demonstrations while the other POMDP RL and IL methods failed. For our future work, we will adapt Seq2Seq IL for closed-loop control, which can lead to better online adaptation. Although we needed to introduce human-in-the-loop learning from expert demonstrations, it also produced substantial boost in sample efficiency, and that opens an intriguing research direction of multi-stage imitation learning.

\bibliographystyle{ieeetr}
\bibliography{example,robotic_insertion_tasks,rl,sim2real,imitation_learning,robot_contact,force_control,Seq2Seq}  %

\begin{thebibliography}{10}

\bibitem{Bekiroglu-2011-ieeetr}
Y.~Bekiroglu, J.~Laaksonen, J.~Jorgensen, V.~Kyrki, and D.~Kragic, ``Assessing
  grasp stability based on learning and haptic data,'' {\em IEEE Transactions
  on Robotics}, vol.~27, no.~3.

\bibitem{Nakagaki-1995-icra}
H.~Nakagaki, K.~Kitagaki, and H.~Tsukune, ``Study of insertion task of a
  flexible beam into a hole,'' in {\em IEEE Int. Conf. on Robotics and
  Automation (ICRA)}, 1999.

\bibitem{Yamamoto-2002-iros}
Y.~Yamamoto, T.~Yoneyama, T.~Hashimoto, T.~Okubo, and T.~Itoh, ``Sensor-based
  analysis of high-precision insertion tasks,'' in {\em IROS}, 2002.

\bibitem{Ma-2021-ieeetii}
Y.~Ma, D.~Xu, and F.~Qin, ``Efficient insertion control for precision assembly
  based on demonstration learning and reinforcement learning,'' {\em IEEE
  Trans. on Industrial Informatics}, 2021.

\bibitem{suarez2016framework}
F.~Su{\'a}rez-Ruiz and Q.-C. Pham, ``A framework for fine robotic assembly,''
  in {\em 2016 IEEE international conference on robotics and automation
  (ICRA)}, pp.~421--426, IEEE, 2016.

\bibitem{argall2009survey}
B.~D. Argall, S.~Chernova, M.~Veloso, and B.~Browning, ``A survey of robot
  learning from demonstration,'' {\em Robotics and autonomous systems},
  vol.~57, no.~5, pp.~469--483, 2009.

\bibitem{calinon2007teacher}
S.~Calinon and A.~G. Billard, ``What is the teacher’s role in robot
  programming by demonstration?: Toward benchmarks for improved learning,''
  {\em Interaction Studies}, vol.~8, no.~3, pp.~441--464, 2007.

\bibitem{calinon2007incremental}
S.~Calinon and A.~Billard, ``Incremental learning of gestures by imitation in a
  humanoid robot,'' in {\em Proceedings of the ACM/IEEE international
  conference on Human-robot interaction}, pp.~255--262, 2007.

\bibitem{billard2016learning}
A.~G. Billard, S.~Calinon, and R.~Dillmann, ``Learning from humans,'' {\em
  Springer handbook of robotics}, pp.~1995--2014, 2016.

\bibitem{racca2016learning}
M.~Racca, J.~Pajarinen, A.~Montebelli, and V.~Kyrki, ``Learning in-contact
  control strategies from demonstration,'' in {\em IEEE/RSJ International
  Conference on Intelligent Robots and Systems (IROS)}, pp.~688--695, IEEE,
  2016.

\bibitem{johannsmeier2019framework}
L.~Johannsmeier, M.~Gerchow, and S.~Haddadin, ``A framework for robot
  manipulation: Skill formalism, meta learning and adaptive control,'' in {\em
  2019 International Conference on Robotics and Automation (ICRA)},
  pp.~5844--5850, IEEE, 2019.

\bibitem{ranjbar2021residual}
A.~Ranjbar, N.~A. Vien, H.~Ziesche, J.~Boedecker, and G.~Neumann, ``Residual
  feedback learning for contact-rich manipulation tasks with uncertainty,'' in
  {\em 2021 IEEE/RSJ International Conference on Intelligent Robots and Systems
  (IROS)}, pp.~2383--2390, IEEE, 2021.

\bibitem{vuong2021learning}
N.~Vuong, H.~Pham, and Q.-C. Pham, ``Learning sequences of manipulation
  primitives for robotic assembly,'' in {\em 2021 IEEE International Conference
  on Robotics and Automation (ICRA)}, pp.~4086--4092, IEEE, 2021.

\bibitem{dong2020compliance}
Y.~Dong, T.~Ren, D.~Wu, and K.~Chen, ``Compliance control for robot
  manipulation in contact with a varied environment based on a new joint torque
  controller,'' {\em Journal of Intelligent \& Robotic Systems}, vol.~99,
  no.~1, pp.~79--90, 2020.

\bibitem{kutsuzawa2018sequence}
K.~Kutsuzawa, S.~Sakaino, and T.~Tsuji, ``Sequence-to-sequence model for
  trajectory planning of nonprehensile manipulation including contact model,''
  {\em IEEE Robotics and Automation Letters}, vol.~3, no.~4, pp.~3606--3613,
  2018.

\bibitem{si2022adaptive}
W.~Si, Y.~Guan, and N.~Wang, ``Adaptive compliant skill learning for
  contact-rich manipulation with human in the loop,'' {\em IEEE Robotics and
  Automation Letters}, vol.~7, no.~3, pp.~5834--5841, 2022.

\bibitem{xu2018feedback}
J.~Xu, Z.~Hou, W.~Wang, B.~Xu, K.~Zhang, and K.~Chen, ``Feedback deep
  deterministic policy gradient with fuzzy reward for robotic multiple
  peg-in-hole assembly tasks,'' {\em IEEE Transactions on Industrial
  Informatics}, vol.~15, no.~3, pp.~1658--1667, 2018.

\bibitem{zhao2021offline}
T.~Z. Zhao, J.~Luo, O.~Sushkov, R.~Pevceviciute, N.~Heess, J.~Scholz,
  S.~Schaal, and S.~Levine, ``Offline meta-reinforcement learning for
  industrial insertion,'' {\em arXiv preprint arXiv:2110.04276}, 2021.

\bibitem{spector2021learning}
O.~Spector and M.~Zacksenhouse, ``Learning contact-rich assembly skills using
  residual admittance policy,'' in {\em 2021 IEEE/RSJ International Conference
  on Intelligent Robots and Systems (IROS)}, pp.~6023--6030, IEEE.

\bibitem{khader2020stability}
S.~A. Khader, H.~Yin, P.~Falco, and D.~Kragic, ``Stability-guaranteed
  reinforcement learning for contact-rich manipulation,'' {\em IEEE Robotics
  and Automation Letters}, vol.~6, no.~1, pp.~1--8, 2020.

\bibitem{levine2014learning}
S.~Levine and P.~Abbeel, ``Learning neural network policies with guided policy
  search under unknown dynamics,'' {\em Advances in neural information
  processing systems}, vol.~27, 2014.

\bibitem{apolinarska2021robotic}
A.~A. Apolinarska, M.~Pacher, H.~Li, N.~Cote, R.~Pastrana, F.~Gramazio, and
  M.~Kohler, ``Robotic assembly of timber joints using reinforcement
  learning,'' {\em Automation in Construction}, vol.~125, p.~103569, 2021.

\bibitem{spector2020deep}
O.~Spector and M.~Zacksenhouse, ``Deep reinforcement learning for contact-rich
  skills using compliant movement primitives,'' {\em arXiv preprint
  arXiv:2008.13223}, 2020.

\bibitem{luo2018deep}
J.~Luo, E.~Solowjow, C.~Wen, J.~A. Ojea, and A.~M. Agogino, ``Deep
  reinforcement learning for robotic assembly of mixed deformable and rigid
  objects,'' in {\em 2018 IEEE/RSJ International Conference on Intelligent
  Robots and Systems (IROS)}, pp.~2062--2069, IEEE, 2018.

\bibitem{bagnell2001solving}
J.~A. Bagnell, A.~Y. Ng, and J.~G. Schneider, ``Solving uncertain markov
  decision processes,'' 2001.

\bibitem{kwon2020inverse}
M.~Kwon, S.~Daptardar, P.~R. Schrater, and X.~Pitkow, ``Inverse rational
  control with partially observable continuous nonlinear dynamics,'' {\em
  Advances in neural information processing systems}, vol.~33, pp.~7898--7909,
  2020.

\bibitem{gangwani2020learning}
T.~Gangwani, J.~Lehman, Q.~Liu, and J.~Peng, ``Learning belief representations
  for imitation learning in pomdps,'' in {\em Uncertainty in Artificial
  Intelligence}, pp.~1061--1071, PMLR, 2020.

\bibitem{arjona2019rudder}
J.~A. Arjona-Medina, M.~Gillhofer, M.~Widrich, T.~Unterthiner, J.~Brandstetter,
  and S.~Hochreiter, ``Rudder: Return decomposition for delayed rewards,'' {\em
  Advances in Neural Information Processing Systems}, vol.~32, 2019.

\bibitem{ren2021learning}
Z.~Ren, R.~Guo, Y.~Zhou, and J.~Peng, ``Learning long-term reward
  redistribution via randomized return decomposition,'' {\em arXiv preprint
  arXiv:2111.13485}, 2021.

\bibitem{meng2021memory}
L.~Meng, R.~Gorbet, and D.~Kuli{\'c}, ``Memory-based deep reinforcement
  learning for pomdps,'' in {\em 2021 IEEE/RSJ International Conference on
  Intelligent Robots and Systems (IROS)}, pp.~5619--5626, IEEE, 2021.

\bibitem{yang2021recurrent}
Z.~Yang and H.~Nguyen, ``Recurrent off-policy baselines for memory-based
  continuous control,'' {\em arXiv preprint arXiv:2110.12628}, 2021.

\bibitem{han2019variational}
D.~Han, K.~Doya, and J.~Tani, ``Variational recurrent models for solving
  partially observable control tasks,'' {\em arXiv preprint arXiv:1912.10703},
  2019.

\bibitem{singh2021structured}
G.~Singh, S.~Peri, J.~Kim, H.~Kim, and S.~Ahn, ``Structured world belief for
  reinforcement learning in pomdp,'' in {\em International Conference on
  Machine Learning}, pp.~9744--9755, PMLR, 2021.

\bibitem{Hausknecht-2015-AAAI}
M.~Hausknecht and P.~Stone, ``Deep recurrent {Q}-learning for partially
  observable {MDPs},'' in {\em AAAI Fall Symposium}, 2015.

\bibitem{Igl2018icml}
M.~Igl, L.~Zintgraf, T.~Le, F.~Wood, and S.~Whiteson, ``Deep variational
  reinforcement learning for {POMDP}s,'' in {\em Int. Conf. on Machine Learning
  (ICML)}, 2018.

\bibitem{Lee-2020-neurips}
A.~Lee, A.~Nagabandi, P.~Abbeel, and S.~Levine, ``Stochastic latent
  actor-critic: Deep reinforcement learning with a latent variable model,'' in
  {\em Conf. on Neural Information Processing Systems (NeurIPS)}, 2020.

\bibitem{igl2018deep}
M.~Igl, L.~Zintgraf, T.~A. Le, F.~Wood, and S.~Whiteson, ``Deep variational
  reinforcement learning for pomdps,'' in {\em International Conference on
  Machine Learning}, pp.~2117--2126, PMLR, 2018.

\bibitem{GAIL}
J.~Ho and S.~Ermon, ``Generative adversarial imitation learning,'' {\em
  Advances in neural information processing systems}, vol.~29, 2016.

\bibitem{IL1}
S.~Desai, I.~Durugkar, H.~Karnan, G.~Warnell, J.~Hanna, and P.~Stone, ``An
  imitation from observation approach to transfer learning with dynamics
  mismatch,'' {\em Advances in Neural Information Processing Systems}, vol.~33,
  pp.~3917--3929, 2020.

\bibitem{IL2}
G.-H. Kim, S.~Seo, J.~Lee, W.~Jeon, H.~Hwang, H.~Yang, and K.-E. Kim,
  ``Demodice: Offline imitation learning with supplementary imperfect
  demonstrations,'' in {\em International Conference on Learning
  Representations}, 2022.

\bibitem{IL3}
R.~Dadashi, L.~Hussenot, M.~Geist, and O.~Pietquin, ``Primal wasserstein
  imitation learning,'' {\em arXiv preprint arXiv:2006.04678}, 2020.

\bibitem{IL4}
W.~Yang, N.~Strokina, J.~Pajarinen, and J.-k. Kamarainen, ``Constrained
  imitation q-learning with earth mover’s distance reward,'' in {\em Deep
  Reinforcement Learning Workshop NeurIPS 2022}, 2022.

\bibitem{IBC}
P.~Florence, C.~Lynch, A.~Zeng, O.~A. Ramirez, A.~Wahid, L.~Downs, A.~Wong,
  J.~Lee, I.~Mordatch, and J.~Tompson, ``Implicit behavioral cloning,'' in {\em
  Conference on Robot Learning}, pp.~158--168, PMLR, 2022.

\bibitem{demodice}
G.-H. Kim, S.~Seo, J.~Lee, W.~Jeon, H.~Hwang, H.~Yang, and K.-E. Kim,
  ``Demodice: Offline imitation learning with supplementary imperfect
  demonstrations,'' in {\em International Conference on Learning
  Representations}, 2022.

\bibitem{garg2019learning}
N.~P. Garg, D.~Hsu, and W.~S. Lee, ``Learning to grasp under uncertainty using
  {POMDPs},'' in {\em International Conference on Robotics and Automation
  (ICRA)}, pp.~2751--2757, IEEE, 2019.

\bibitem{Montana-1988-ijrr}
D.~Montana, ``The kinematics of contact and grasp,'' {\em The International
  Journal of Robotics Research}, vol.~7, no.~3, 1988.

\bibitem{Howard-1998-mmt}
S.~Howard, M.~Zefran, and V.~Kumar, ``On the $6\times 6$ cartesian stiffness
  matrix for three-dimensional motions,'' {\em Mechanism and Machine Theory},
  vol.~33, no.~4, 1998.

\bibitem{Gold-2020-iros}
T.~Gold, A.~Volz, and K.~Graichen, ``Model {Predictive} {Position} and {Force}
  {Trajectory} {Tracking} {Control} for {Robot}-{Environment} {Interaction},''
  in {\em {IEEE}/{RSJ} {International} {Conference} on {Intelligent} {Robots}
  and {Systems} ({IROS})}, 2020.

\bibitem{kaelbling1998planning}
L.~P. Kaelbling, M.~L. Littman, and A.~R. Cassandra, ``Planning and acting in
  partially observable stochastic domains,'' {\em Artificial Intelligence},
  vol.~101, no.~1-2, pp.~99--134, 1998.

\bibitem{nguyen2020belief}
H.~Nguyen, B.~Daley, X.~Song, C.~Amato, and R.~Platt, ``Belief-grounded
  networks for accelerated robot learning under partial observability,'' {\em
  arXiv preprint arXiv:2010.09170}, 2020.

\bibitem{rajeswar2022haptics}
S.~Rajeswar, C.~Ibrahim, N.~Surya, F.~Golemo, D.~Vazquez, A.~Courville, and
  P.~O. Pinheiro, ``Haptics-based curiosity for sparse-reward tasks,'' in {\em
  Conference on Robot Learning}, pp.~395--405, PMLR, 2022.

\bibitem{sutskever2014sequence}
I.~Sutskever, O.~Vinyals, and Q.~V. Le, ``Sequence to sequence learning with
  neural networks,'' {\em Advances in neural information processing systems},
  vol.~27, 2014.

\bibitem{brown2020language}
T.~Brown, B.~Mann, N.~Ryder, M.~Subbiah, J.~D. Kaplan, P.~Dhariwal,
  A.~Neelakantan, P.~Shyam, G.~Sastry, A.~Askell, {\em et~al.}, ``Language
  models are few-shot learners,'' {\em Advances in neural information
  processing systems}, vol.~33, pp.~1877--1901, 2020.

\bibitem{ross2011reduction}
S.~Ross, G.~Gordon, and D.~Bagnell, ``A reduction of imitation learning and
  structured prediction to no-regret online learning,'' in {\em Proceedings of
  the fourteenth international conference on artificial intelligence and
  statistics}, pp.~627--635, JMLR Workshop and Conference Proceedings, 2011.

\bibitem{ni2021recurrent}
T.~Ni, B.~Eysenbach, and R.~Salakhutdinov, ``Recurrent model-free rl can be a
  strong baseline for many pomdps,'' in {\em Int. Conf. on Machine Learning
  (ICML)}, 2022.

\bibitem{nair2018overcoming}
A.~Nair, B.~McGrew, M.~Andrychowicz, W.~Zaremba, and P.~Abbeel, ``Overcoming
  exploration in reinforcement learning with demonstrations,'' in {\em 2018
  IEEE international conference on robotics and automation (ICRA)},
  pp.~6292--6299, IEEE, 2018.

\bibitem{haarnoja2018soft}
T.~Haarnoja, A.~Zhou, P.~Abbeel, and S.~Levine, ``Soft actor-critic: Off-policy
  maximum entropy deep reinforcement learning with a stochastic actor,'' in
  {\em International conference on machine learning}, pp.~1861--1870, PMLR,
  2018.

\bibitem{reddy2019sqil}
S.~Reddy, A.~D. Dragan, and S.~Levine, ``Sqil: Imitation learning via
  reinforcement learning with sparse rewards,'' {\em arXiv preprint
  arXiv:1905.11108}, 2019.

\bibitem{bain1995framework}
M.~Bain and C.~Sammut, ``A framework for behavioural cloning.,'' in {\em
  Machine Intelligence 15}, pp.~103--129, 1995.

\end{thebibliography}

\end{document}